\documentclass[A4paper, 11 pt, conference]{ieeeconf}  % Comment this line out if you need a4paper

\IEEEoverridecommandlockouts                              % This command is only needed if 
\usepackage{geometry}
\usepackage{graphics} % for pdf, bitmapped graphics files
\usepackage{amsmath} % assumes amsmath package installed
\usepackage{amssymb}  % assumes amsmath package installed
\usepackage{xcolor}
\usepackage{epsfig}
\usepackage{graphicx}
\usepackage{subfigure}
\usepackage{epstopdf} % for eps figure using pdflatex
\usepackage[linesnumbered,ruled,vlined]{algorithm2e}

\usepackage[utf8]{inputenc}

\title{\LARGE \bf
Learning-based Fast Path Planning in Complex Environments
}

\author{Jianbang Liu$^1$, Baopu Li$^2$, Tingguang Li$^3$, Wenzheng Chi$^4$,\\ Jiankun Wang$^{5*}$ and Max Q.-H. Meng$^{56*}$, \emph{Fellow, IEEE}% <-this % stops a space
	\thanks{*This work is partially supported by National Key R\&D program of China with Grant No. 2019YFB1312400, Shenzhen Key Laboratory of Robotics Perception and Intelligence (ZDSYS20200810171800001), GRF grant \#14200618 and Hong Kong RGC CRF grant C4063-18G.(\emph{Corresponding authors: Jiankun Wang, Max Q.-H. Meng})}
	\thanks{$^1$Department of Electronic Engineering, The Chinese University of Hong Kong, Hong Kong, henryliu@link.cuhk.edu.hk.}%
	\thanks{$^2$Baidu Research (US), CA, USA, bpli.cuhk@gmail.com.}%
	\thanks{$^3$Tencent Robotics X, Shenzhen, China, teaganli@tencent.com.}%
	\thanks{$^4$School of Mechanical and Electric Engineering, Soochow University, Suzhou, China, wzchi@suda.edu.cn.}%
	\thanks{$^5$Department of Electronic and Electrical Engineering of the Southern University of Science and Technology, Shenzhen, China, \{wangjk,mengqh\}@sustech.edu.cn.}
	\thanks{$^6$Max Q.-H. Meng is on leave from the Department of Electronic Engineering, The Chinese University of Hong Kong, Hong Kong, and also with the Shenzhen Research Institute of the Chinese University of Hong Kong, Shenzhen, China.}
}

\begin{document}

\maketitle
\thispagestyle{empty}
\pagestyle{empty}

%%%%%%%%%%%%%%%%%%%%%%%%%%%%%%%%%%%%%%%%%%%%%%%%%%%%%%%%%%%%%%%%%%%%%%%%%%%%%%%%
\begin{abstract}
In this paper, we present a novel path planning algorithm to achieve fast path planning in complex environments.
Most existing path planning algorithms are difficult to quickly find a feasible path in complex environments or even fail.
However, our proposed framework can overcome this difficulty by using a learning-based prediction module and a sampling-based path planning module.
The prediction module utilizes an auto-encoder-decoder-like convolutional neural network (CNN) to output a promising region where the feasible path probably lies in.
In this process, the environment is treated as RGB image to feed in our designed CNN module, and the output is also RGB image.
No extra computation is required so that we can maintain a high processing speed of $60$ frame-per-second (FPS).
Incorporated with a sampling-based path planner, we can extract a feasible path from the output image so that the robot can track it from start to goal.
To demonstrate the advantage of the proposed algorithm, we compare it with conventional path planning algorithms in a series of simulation experiments.
The results reveal that the proposed algorithm can achieve much better performance in terms of planning time, success rate, and path length.
\end{abstract}

%%%%%%%%%%%%%%%%%%%%%%%%%%%%%%%%%%%%%%%%%%%%%%%%%%%%%%%%%%%%%%%%%%%%%%%%%%%%%%%%
\section{INTRODUCTION}
\label{Introduction}

\begin{figure}[t]
	\centering
	\subfigure[RRT* in 100s.]{
		\includegraphics[width=39mm]{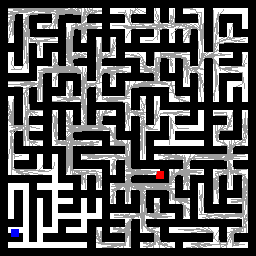}}
	\subfigure[Our algorithm in 25s.]{
		\includegraphics[width=39mm]{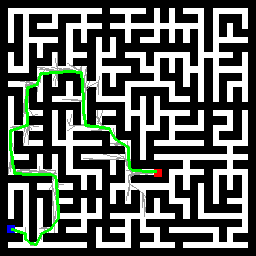}}
	\caption{The red and blue rectangles represent the start and goal state, respectively. The black and white area represent the obstacle and free space, the green line represents the feasible path connecting the start and goal, and the gray lines represent the generated random tree, respectively. (a) The RRT* cannot find a feasible path within limited time (100s). (b) The proposed learning-based RRT* can find a feasible path within 25s.}
	\label{fig_intro}
	\vspace{-4mm}
\end{figure}

Path planning is a fundamental problem in robotics \cite{siciliano2016springer}, which aims to generate a collision-free path to drive the robot to move from start position to goal.
Sampling-based path planning algorithms have been shown to be successful for many path planning problems.
With a collision detection module to determine whether a single state is collision-free, they can avoid the complex geometric modeling of the configuration space.
By applying the collision detector to different robots and environments, sampling-based planners can be used to solve a broad class of path planning problems.
Two representative algorithms are rapidly-exploring random tree (RRT) \cite{lavalle2001randomized} and probabilistic roadmap (PRM) \cite{kavraki1996probabilistic}.
Many variants are also proposed, which either rely on heuristic sampling techniques \cite{wang2020optimal}\cite{gammell2014informed} or combine with certain theories for specific applications \cite{wang2019socially}.
However, the sampling-based algorithm only guarantees a weaker form of completeness.
As the number of iterations goes to infinity, the planner will eventually find a solution if it exists.
This is because the sampling-based planner implements a probabilistic sampling method.
It means that samples in the entire state space are selected using a uniform probability density function.
This uniform sampling strategy indeed guarantees the planner's probabilistic completeness.
However, it performs badly or even fails in complex environments such as the maze environment.
As shown in Fig. \ref{fig_intro}, the sampling-based planner cannot find a feasible path within the limited time.

\begin{figure}[t]
	\centering
	\includegraphics[width=80mm]{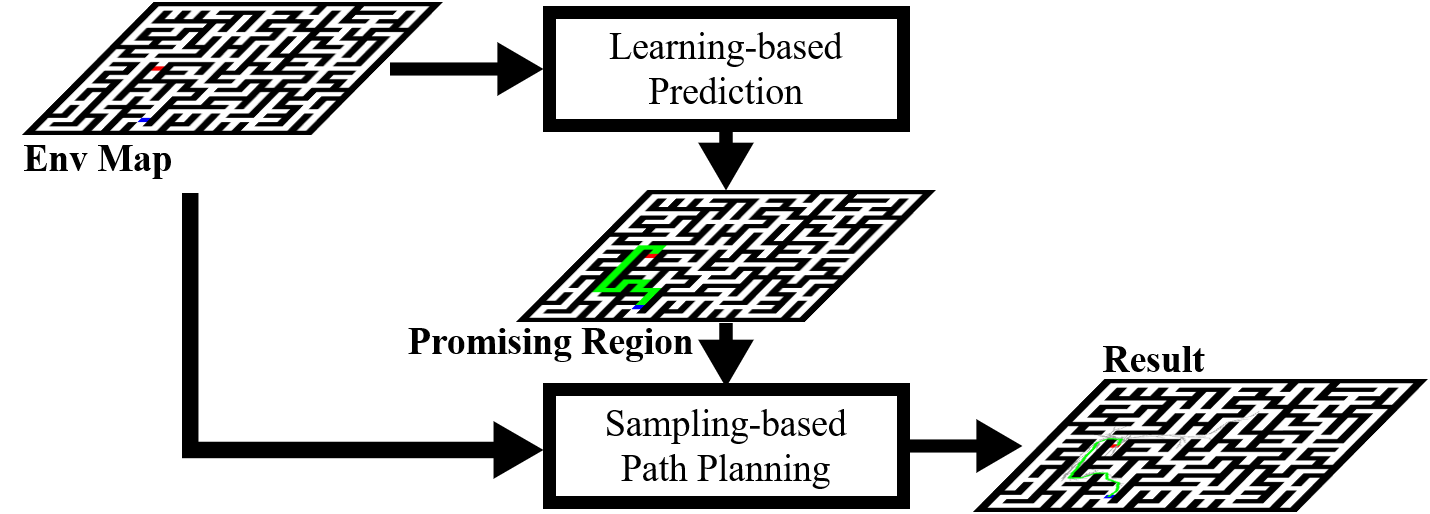}
	\caption{Framework of the proposed learning-based RRT*.}
	\label{fig_pipeline}
\end{figure}

To address this problem, we present a novel path planning framework to achieve fast path planning in complex environments.
Unlike conventional methods in which a human-crafted or environment-related heuristic is designed for specific environments, our proposed algorithm provides a more general solution.
The pipeline of how it solves the path planning problem is illustrated in Fig. \ref{fig_pipeline}.
As we can see, the proposed framework mainly consists of two parts: a learning-based prediction module and a sampling-based path planning module.
In the prediction module, a convolutional neural network (CNN) is taken to preprocess the given environment information.
We denote the environment as a RGB image where the start, goal, free space, and obstacle space are represented with different colors, respectively.
It means that we do not need to access the geometric structure of a given environment.
Instead, we process the environment information at an image level.
The output of the prediction model is also an image.
The promising region where a feasible path probably lies is highlighted.
In the sampling-based path planning module, a RRT* planner is used to generate a feasible path based on the promising region.
With this generated path, the robot can track it from start to goal.

Our contributions are summarized as follows:
\begin{itemize}
	\item A novel sampling method for fast path generation in complex environments;
	\item An efficient neural network to predict the promising region for the given complex environment;
	\item A series of case studies to demonstrate the advantage of the proposed algorithm.
\end{itemize}

The rest of this paper is organized as follows.
We first review the related work in Section \ref{sec_related}.
Section \ref{sec_alg} introduces the details of our proposed path planning algorithm for complex environments.
Then simulation results are reported and analyzed in Section \ref{sec_result}.
Finally, we conclude this paper and discuss future work in Section \ref{sec_conclusion}.

\section{Related Work}
\label{sec_related}

Sampling-based path planning algorithms are very popular because they can efficiently search the state space.
However, they perform poorly in certain cases, especially in environments with narrow passages and bug traps.
To overcome these limitations, many algorithms have been proposed and most of them rely on biased sampling or sample rejection.
Lee \emph{et al.} \cite{lee2014local} propose a local path planning method for self-driving cars in a complex environment, where a novel path representation and modification method based on Voronoi cell is implemented.
Liu \emph{et al.} \cite{liu2019evolution} use an evolution optimization method to achieve path planning in complex environments, which performs like the artificial potential filed (APF) method \cite{khatib1986real}.
A disadvantage is that this method cannot work well in environments with some turns or bug traps.
In \cite{wang2019finding}, Wang \emph{et al.} introduce a Gaussian mixture model to quickly generate a high-quality path, but the parameters are required to be tuned for different environments.
These aforementioned methods may work well in their proposed scenarios but cannot be generalized to other environments.

There are also some sampling-based algorithms that are applicable to different complex environments.
They usually use a graph search theory to provide prior knowledge of the current environment.
A*-RRT* algorithm \cite{brunner2013hierarchical} uses the A* search method \cite{hart1968formal} to guide the sampling process.
However, the computation cost of the A* algorithm exponentially increases as the problem scale becomes bigger since it requires the discretization of the current environment.
Potentially guided bidirectional RRT* (PB-RRT*) \cite{tahir2018potentially} uses the APF method as a heuristic to accelerate the path planning process, but this heuristic cannot provide a reasonable bias to a feasible path in complex environments.

\begin{figure*}[t]
	\centering
	\includegraphics[width=172mm]{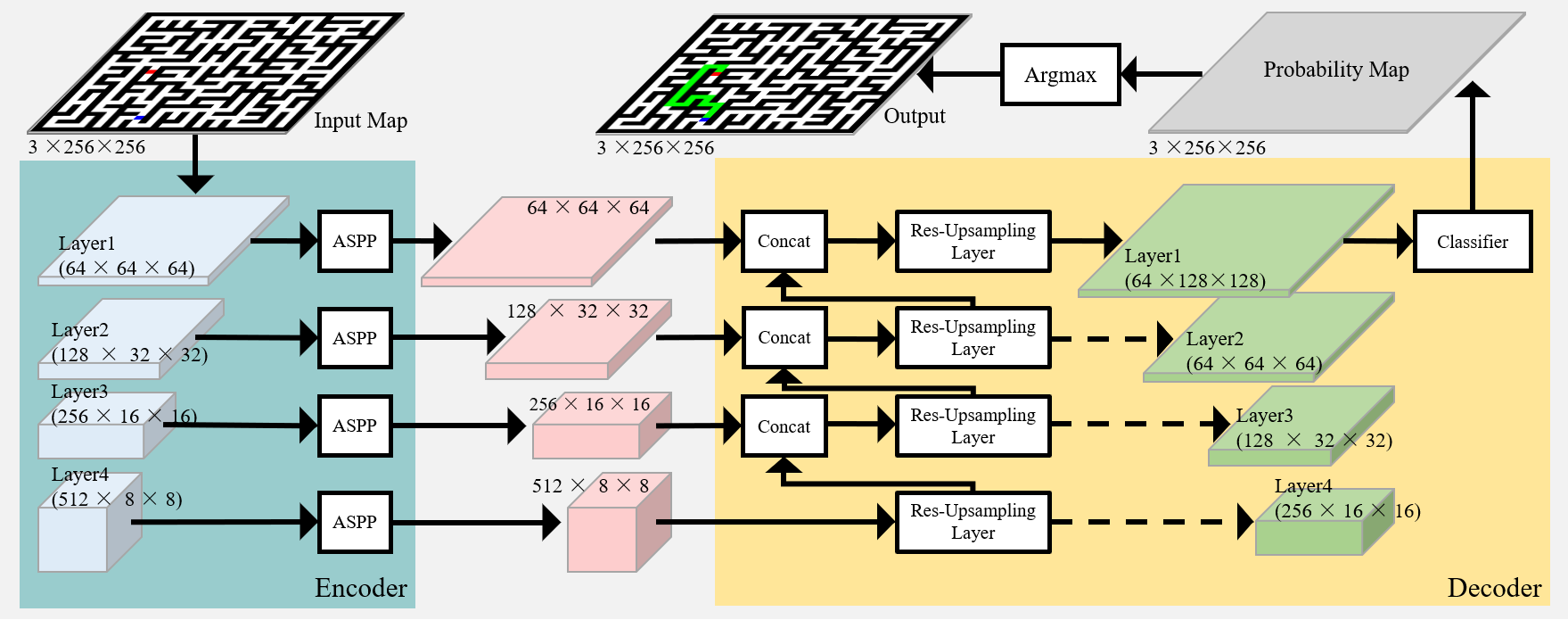}
	\caption{Illustration of network architecture. ASPP denotes the atrous pyramid pooling. The teal blue part denotes the encoder structure and the yellow part denotes the decoder structure, respectively.}
	\label{fig_arch}
	\vspace{-4mm}
\end{figure*}

% Recently, learning-based methods are also widely applied to the path planning field.
Recently, learning-based methods are widely applied into robotic tasks, such as grasping \cite{9197256}, autonomous driving \cite{jenni2020steering}, and robot-assisted surgery \cite{9196560}.
Among the literatures addressing the path planning, Qureshi \emph{et al.} \cite{qureshi2019motion} propose the motion planning network (MPN) to generate an end-to-end feasible path from the point cloud.
Zhang \emph{et al.} \cite{zhang2018learning} implement a policy-based search method to learn an implicit sampling distribution for specific environments.
De \emph{et al.} \cite{de2019learning} propose to learn a lattice planner control set to achieve path planning for autonomous vehicles.
The imitation learning methods such as conditional variational auto-encoder (CVAE) \cite{ichter2018learning}, generative adversarial network (GAN) \cite{zhang2020generative} and recurrent neural network (RNN) \cite{li2020efficient} are used to bias the search direction through various probabilistic models.
However, specially designed local features or parametrized data structure is required in the learning process, which limits the generalization ability.

In this paper, our proposed learning-based path planning algorithm does not need to discretize the state space or design specific local features.
It directly uses the RGB image of the current environment as input, where start, goal, free space, and obstacle space are denoted with different colors, respectively.
Then the output is treated as a heuristic to guide a RRT* planner to achieve fast path planning.
Our proposed method is applicable to different environments and can also be easily extended to other sampling-based planners.

\section{Algorithm}
\label{sec_alg}
Inspired by the previous work \cite{wang2020neural}, we propose an auto-encoder-decoder-like CNN structure to achieve promising region prediction so that the performance of the path planning algorithm can be improved significantly.
Herein, a light-weighted and powerful network structure is designed to deal with complex environments, which learns from the training data (pairs of the map and ground truth).
When the training process is done for a given map (represented as RGB image), our designed network can compute the probabilities of every pixel being classified as certain kinds of map elements, such as free space, obstacle space, or promising region.
The generated promising region is used to guide the sampling process of the path planner, resulting in an efficient search of the state space.
Therefore, the performance of the sampling-based path planner is naturally improved. 

\subsection{Network Structure}
\label{sec_alg_struct}

In complex environments, it is challenging for the sampling-based path planner to find a feasible path since it employs a uniform sampling technique, and this technique is sensitive to spatial information.
Therefore, the network should capture the characteristic of spatial information and provide effective guidance for the path planner.
Nevertheless, the general encoder in CNN gradually decreases the spatial resolution of feature maps, which only reserves partial contextual information and omits the spatial details.
To address this problem, we propose a novel decoder to reconstruct the environment map and locate the promising region in a coarse-to-fine manner.
On the one hand, the encoder extracts the multi-resolution feature maps and delivers them to the decoder.
On the other hand, the decoder fuses the multi-resolution feature maps layer by layer.
Finally, the classifier produces a score map indicating the probabilities of each pixel being classified as a specified class.
The overall structure of our designed neural network is shown in Fig. \ref{fig_arch}.

\subsubsection{Encoder}
The encoder is fed with RGB images that denote the environment map, start, and goal to generate a high-dimensional representation of the environment map.
The ResNet$18$ is deployed as the encoder to extract multi-resolution feature maps.
The encoder is divided into $4$ layers, and each layer extracts corresponding feature maps with a specified resolution.
The current layer transforms the feature maps generated from the previous layer to a higher dimensional representation, and reduces the resolution of the feature maps by a factor of $2$.
The pyramid pooling has been empirically proven to be effective in reserving contextual information with various scale \cite{zhao2017pyramid}\cite{chen2017rethinking}\cite{wu2019fastfcn}.
Considering the diversity of map elements in shape and scale within complex environments, we implement the atrous pyramid pooling module (ASPP) \cite{chen2017rethinking} in our network.
The ASPP module can detect features across scales and prevent the network from being affected by the small alteration of map complexity, such as changing the scale of obstacle or narrow passage.

\begin{figure}[t]
	\centering
	\includegraphics[width=50mm]{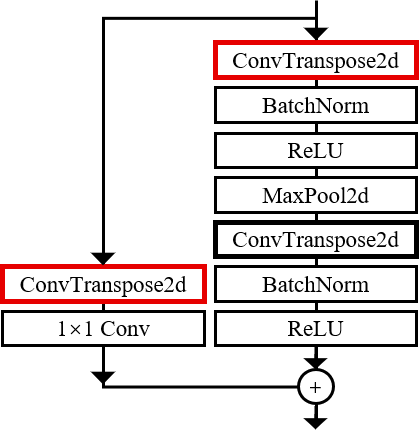}
	\caption{The proposed structure of residual up-sampling block. The two ConvTranspose2D layers in red can either enlarge the spatial resolution of input or maintain the resolution. The ConvTransposed2D layer in black keeps the resolution unchanged and makes the block deeper.}
	\label{fig_blk_struct}
	\vspace{-4mm}
\end{figure}

\subsubsection{Decoder}
A residual decoder block is constructed by replacing the convolution layer in the residual block of ResNet$18$ with the deconvolution layer.
The structure of the residual decoder block is shown in Fig. \ref{fig_blk_struct}.
The residual decoder block can up-sample or maintain the spatial resolution of the feature maps.
To up-sample the resolution of the feature maps, the kernel size of the leading deconvolution layer.
The kernel size of the deconvolution layer in the residual path will also be set to $4$.
The stride of these two deconvolution layers will be set to $2$. 
These two deconvolution layers are highlighted in red in Fig. \ref{fig_blk_struct}.
If the block is designed to maintain the resolution, the kernel size and stride of all deconvolution layers in the decoder block will be configured as $3$ and $1$ accordingly.
Each residual up-sampling layer consists of two serial connected residual decoder blocks.
The up-sampled feature map will be concatenated with the feature maps with the same resolution generated from the encoder. 
Then, the stacked feature maps are passed to the next up-sampling layer.
The final up-sampling layer will fully recover the resolution of the feature map to that of the input map. 
In the end, a single convolution layer serves as the classifier to generate the probability map.
This classifier predicts the probability of each pixel being classified as the predefined classes. 

\subsection{Loss and Evaluation}
\label{sec_alg_train_eval}
A weighted focal loss \cite{lin2017focal} is implemented during the network training, which enforces to pay more attention to the challenging classification cases.
The weight is set to eliminate the imbalance in the total number of pixels among different classes so that the network will not be restricted by the overall frequency of a pixel classified into a certain class.
This is because the number of pixels in free space is much more than the pixels belonging to the promising region.
The loss function can be formulated as follows:

\begin{equation} \label{fn_loss}
\begin{split}
l_{i,j} & = \\
& -\sum_{k}^{N-1}w^k g^k_{i,j}[1-S(p^k_{i,j})^\gamma]log[S(p^k_{i,j})],
\end{split}
\end{equation}
where $(i,j)$ indicates the position of the pixel on the map and $k$ indicates index of the class among total $N$ classes.
$p^k_{i,j}$ represents the predicted probability of the pixel at $(i,j)$ being classified as the $k$th class.
$g$ represents the ground truth and
\begin{equation}
g^k_{i,j} = \begin{cases} 
1 & \text{if}\; (i,j)\; \text{belongs to the $k$th class} \\
0 & \text{else}
\end{cases}.
\end{equation}
$\gamma$ is the focusing parameter and $S(\cdot)$ stands for the soft-max function.
$w^k$ represents the weight assigned to $k$th class to handle the unbalance issue of data.

Accordingly, we propose a novel metric to evaluate the performance of network model:

\begin{equation} \label{fn_metric}
\begin{split}
met & ric =  \\ 
& 1 - \frac{ \displaystyle \sum_{i=0}^{H-1} \sum_{j=0}^{W-1} c_{i,j} \cdot (g^{pr}_{i,j} - g^{free}_{i,j}) }{ \displaystyle \sum_{i=0}^{H-1} \sum_{j=0}^{W-1} g^{pr}_{i,j} }
\end{split}
\end{equation}
\noindent where $c$ represents the classification result and
\begin{equation} \label{fn_c_pr}
c_{i,j} = 
\begin{cases}
1 &if\; (i,j) \in Promising\; Region \\
0 &else
\end{cases},
\end{equation}

\noindent $g$ represents the ground truth and
\begin{equation} \label{fn_gt_pr} 
g^{pr}_{i,j} = 
\begin{cases}
1 &if\; (i,j) \in Promising\; Region \\
0 &else
\end{cases}, 
\end{equation}

\begin{equation} \label{fn_gt_free} 
g^{free}_{i,j} = 
\begin{cases}
1 &if\; (i,j) \in free space\\
0 &else
\end{cases}.
\end{equation}

In the metric, $\displaystyle \sum_{i=0}^{H-1} \sum_{j=0}^{W-1} c_{i,j} g^{pr}_{i,j}$ counts the number of correctly classified pixels belonging to the promising region.
The accuracy of the prediction is defined by the percentage of the promising region pixels in ground truth correctly labeled in prediction.

\begin{equation}
Accuracy = \frac{\displaystyle \sum_{i=0}^{H-1} \sum_{j=0}^{W-1} c_{i,j}g^{pr}_{i,j}}{\displaystyle \sum_{i=0}^{H-1} \sum_{j=0}^{W-1} g^{pr}_{i,j}}
\end{equation}

$\displaystyle \sum_{i=0}^{H-1} \sum_{j=0}^{W-1} c_{i,j}g^{free}_{i,j}$ counts the number of pixels belonging to free space in ground truth that are labeled as promising region in evaluation.

The redundancy is defined by the the ratio of $\displaystyle \sum_{i=0}^{H-1} \sum_{j=0}^{W-1} c_{i,j}g^{free}_{i,j}$ to the number of pixels belonging to promising region in ground truth. 
\begin{equation}
Redundancy = \frac{\displaystyle \sum_{i=0}^{H-1} \sum_{j=0}^{W-1} c_{i,j}g^{free}_{i,j}}{\displaystyle \sum_{i=0}^{H-1} \sum_{j=0}^{W-1} g^{pr}_{i,j}}
\end{equation}

Thus, this metric is formed by merging the accuracy and the redundancy of the prediction result:
\begin{equation}
metric = (1 - Accuracy) + Redundancy,
\end{equation}
a lower value of the metric suggests a better performance in the evaluation.

\begin{figure}[t]
	\vspace{-4mm}
	\centering
	\subfigure[Example of map image.]{
		\includegraphics[width=39mm]{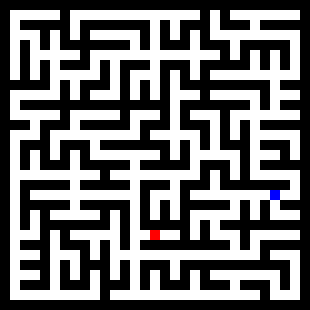}}
	\subfigure[Example of ground truth image.]{
		\includegraphics[width=39mm]{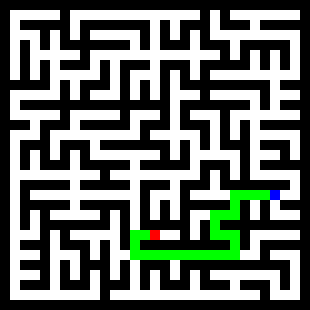}}
	\caption{An example of samples with 31$\times$31 map complexity: The block in red and blue represent the start and goal point. The map blocks in black represent obstacles, the blocks in white represent free space, and the map blocks in green represent the feasible path connecting the start and goal.}
	\label{fig_example_training_sample}
\end{figure}

\section{Simulation Experiment Results}
\label{sec_result}
A series of simulation experiments have been conducted with the maze environment, which is pretty arduous for the sampling-based path planner to find a feasible path. 
The maze map is composited by $2$ kinds of block, free block (free space), and obstacle block (obstacle space). 
A block can occupy several $n \times n$ pixels on the map image, and the size of blocks is consistent over the whole map.
Each map consists of $m \times m$ blocks.
The complexity of the maze varies by changing the number of blocks in the column and row.
An example of $31 \times 31$ maze map is shown in Fig. \ref{fig_example_training_sample}.
We construct a dataset containing different sizes of maze maps from $21 \times 21$ to $49 \times 49$. 
For each complexity level, $8000$ samples are collected, in which $6000$ samples are reserved for training, $1000$ samples for evaluation and the remaining for test.

\def\sqPDF#1#2{0 0 m #1 0 l #1 #1 l 0 #1 l h}
\def\circPDF#1#2{#1 0 0 #1 #2 #2 cm .1 w .5 0 m
	.5 .276 .276 .5 0 .5 c -.276 .5 -.5 .276 -.5 0 c
	-.5 -.276 -.276 -.5 0 -.5 c .276 -.5 .5 -.276 .5 0 c h}
\def\diaPDF#1#2{#2 0 m #1 #2 l #2 #1 l 0 #2 l h}

\def\credCOLOR   {.54 .14 0}
\def\cblueCOLOR  {.06 .3 .54}
\def\cgreenCOLOR {0 .66 0}
\def\cblackCOLOR {0 0 0}

\def\genbox#1#2#3#4#5#6{
	% #1=0/1, #2=color, #3=shape, #4=raise, #5=width, #6=width/2
	\leavevmode\raise#4bp\hbox to#5bp{\vrule height#5bp depth0bp width0bp
		\pdfliteral{q .5 w \csname #2COLOR\endcsname\space RG
			\csname #3PDF\endcsname{#5}{#6} S Q
			\ifx1#1 q \csname #2COLOR\endcsname\space rg 
			\csname #3PDF\endcsname{#5}{#6} f Q\fi} \hss}}

% shape     raise width  width/2
\def\sqbox      #1#2{\genbox{#1}{#2}  {sq}       {0}   {4.5}  {2.25}}
\def\circbox    #1#2{\genbox{#1}{#2}  {circ}     {0}   {5}    {2.5}}
\def\diabox     #1#2{\genbox{#1}{#2}  {dia}      {-.5} {6}    {3}}

\begin{algorithm}[b]
	\DontPrintSemicolon
	\SetKwRepeat{Do}{do}{while}
	\SetKwInOut{Input}{Input}
	\SetKwInOut{Output}{Output}
	\Input{$x_{init}, \mathcal{X}_{goal} \ and \ Map$}
	\Output{$\mathcal{T}$}
	$\mathcal{V} \leftarrow x_{init}, \mathcal{E} \leftarrow \emptyset, \mathcal{T} = (\mathcal{V},\mathcal{E})$;\;
	\For{$i = 1...N$}{
		\If{Rand($0,1$) $ < \alpha $}{
			$x_{rand} \leftarrow \text{LearningBasedSampling}()$;\;}
		\Else{
			$x_{rand} \leftarrow \text{UniformSampling}()$;\;}
		$x_{nearest} \leftarrow \text{Nearest}(\mathcal{T}, x_{rand})$;\;
		$x_{new} \leftarrow \text{Steer}(x_{nearest},x_{rand})$;\;
		\If{$\text{ObstacleFree}(x_{nearest},x_{new})$}{
			$\mathcal{T} = \text{Extend}(\mathcal{T}, x_{new})$;\;
			$\text{Rewire()}$;\;
			\If{$x_{new} \in \mathcal{X}_{goal}$}{
				$\text{Return} \ \mathcal{T}$;\;
			}
		}
	}
	$\text{Return} \ failure$;\;
	\caption{L-RRT*. \label{Algs.L-RRT*}}
\end{algorithm}

\subsection{Promising Region Prediction}
\label{sec_result_pred}
We train our prediction network with samples from $3$ complexity levels,  $31 \times 31$, $33 \times 33$ and $35 \times 35$. 
In this experiment, we deploy the Adam optimizer with $0.001$ as the initial learning rate. 
The models are trained for $30$ epochs on a desktop with Intel(R) Core(TM) i7-10700K CPU @ 3.80GHz, 64G RAM and 2 NVIDIA RTX 2080 GPUs. 
The inference experiments are conducted with only one NVIDIA RTX 2080 GPU.

\begin{figure}[t]
	\centering
	\includegraphics[width=80mm]{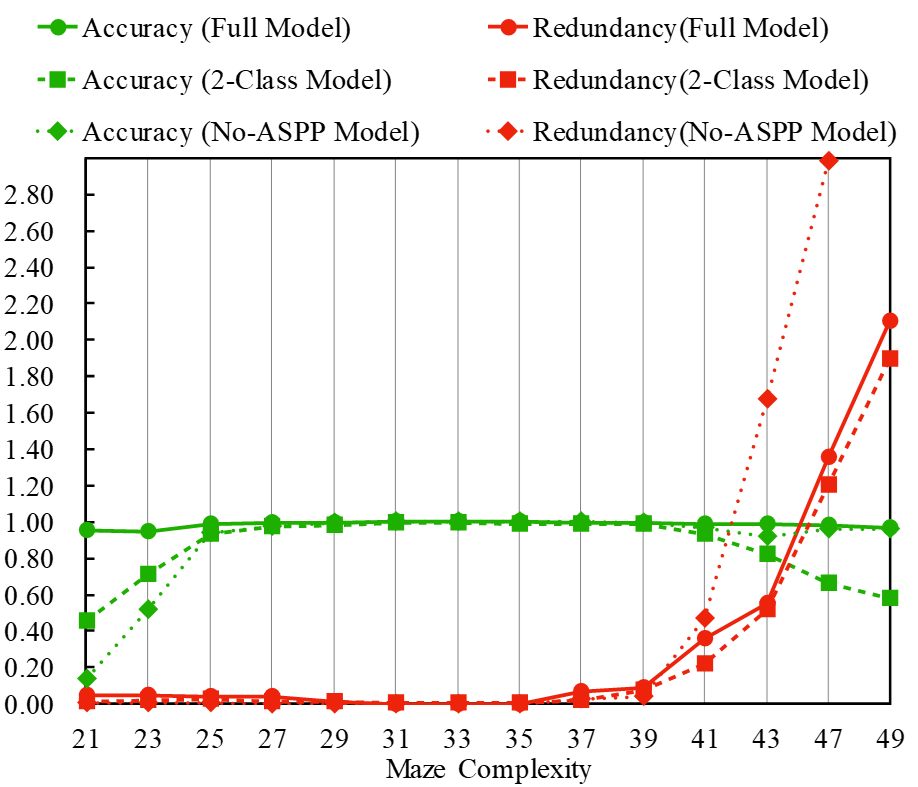}
	\caption{The evaluation result of (1) full model, (2) 2-class model and (3) model without ASPP.}
	\label{fig_eval}
	\vspace{-4mm}
\end{figure}

In our full model, the pixels are classified into $3$ classes, free space, promising region, and obstacle space to strengthen the awareness of the network on the details of obstacles by such explicit supervision.
As shown in Fig. \ref{fig_eval} that includes the evaluations on all the test samples, the mean accuracy and redundancy of full model's performance in processing maze map with complexity from $21$ to $49$ is denoted by the green and red symbol ``\circbox1{cblack}''.
A $2$-class model is also trained for comparison. 
Its accuracy and redundancy are denoted by the green and red symbol ``\sqbox1{cblack}''.
A model without the ASPP module is trained to validate that the coarse-to-fine up-sampling structure acquires the ability to handle environments with unseen complexity. 
Then, the ASPP module further boosts the generalization ability of the designed network against change in scale.
The green and red symbol ``\diabox1{cblack}'' denotes the accuracy and redundancy of the prediction of the model without the ASPP module.
%The evaluation result on the all test samples in the dataset is shown in Fig. \ref{fig_eval}, where the accuracy and redundancy are shown by curves with different colors(the accuracy is shown in green and the redundancy is shown in red).

The full model achieves high accuracy across all complexity levels.
Explicitly involving obstacle class is shown to be beneficial since the full model performs better than the $2$-class model in maze map with lower unseen complexity.
When encountering higher unseen complexity, the no-ASPP model maintains the accuracy by largely raising the redundancy in the prediction, which makes the prediction less instructive.
Besides, our model can make inference on an NVIDIA RTX 2080 GPU with over 60 FPS, which makes the real-time promising region prediction possible in autonomous systems.

\begin{figure*}[t]
	\centering
	\subfigure[]{
		\includegraphics[width=45mm]{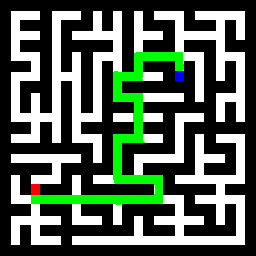}
		\label{fig_text_result_25_env}}
	\subfigure[]{
		\includegraphics[width=45mm]{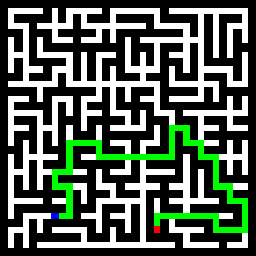}
		\label{fig_text_result_35_env}}
	\subfigure[]{
		\includegraphics[width=45mm]{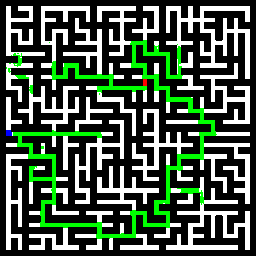}
		\label{fig_test_result_45_env}}
	\caption{The selected environments for experiments:
		(a) $25 \times 25$ test environment with start, goal and predicted promising region; 
		(b) $35 \times 35$ test environment with start, goal and predicted promising region;
		(c) $45 \times 45$ test environment with start, goal and predicted promising region.}
	\label{fig_test_envs}
\end{figure*}

\begin{figure*}[t]
	\centering
	\subfigure[]{
		\includegraphics[width=52mm]{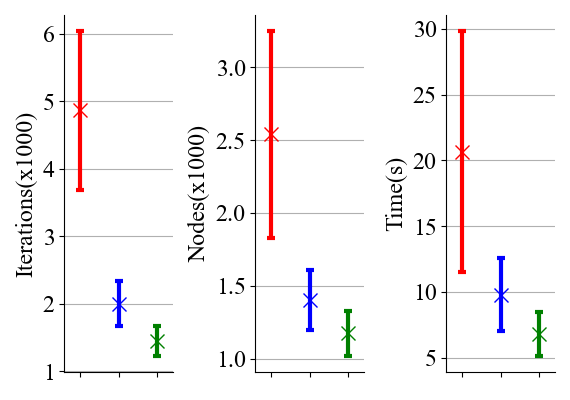}
		\label{fig_test_result_25_stat}}
	\subfigure[]{
		\includegraphics[width=52mm]{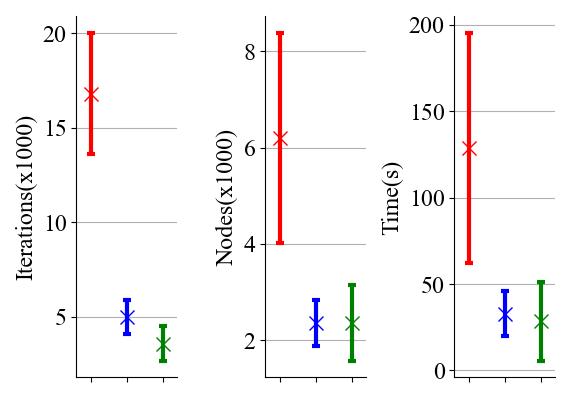}
		\label{fig_test_result_35_stat}}
	\subfigure[]{
		\includegraphics[width=52mm]{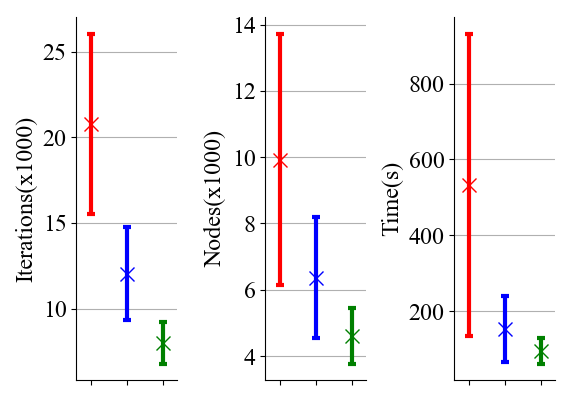}
		\label{fig_test_result_45_stat}}
	\caption{The experiment results of the algorithms in the selected environments. The cross mark denotes the mean and the line denotes the standard deviation. The results of conventional RRT* algorithm are represented in red color, the results of L-RRT*($\alpha=0.5$) are represented in blue color and the results of L-RRT*($\alpha=0.8$) are represented in green color:
		(a) Test results in $25 \times 25$ maze;
		(b) Test results in $35 \times 35$ maze;
		(c) Test results in $45 \times 45$ maze.}
	\label{fig_test_result}
	\vspace{-4mm}
\end{figure*}

\subsection{Path Planning Results}
\label{sec_planning_result}
The promising region prediction is combined with the RRT* algorithm to plan a feasible path from the start to the goal in the given maze environment.
The designed RRT* algorithm utilizing the learning-based promising region prediction is denoted as L-RRT*.
Apart from the conventional RRT* algorithm which employs a uniform sampler, L-RRT* implements two samplers, the uniform sampler and the biased sampler.
The biased sampler will randomly sample nodes on the predicted promising region, while the uniform sampler samples nodes on the entire maze map.
The probability of using the biased sampler is controlled by a factor $\alpha \in (0, 1]$. When the $\alpha$ is $0.5$, the probability of using the biased sampler is $50\%$.
Except for the sampling scheme, the rest part of the L-RRT* is the same as the conventional RRT*.
The detail of our L-RRT* is shown in Alg. \ref{Algs.L-RRT*}.

Our L-RRT* is compared with the conventional RRT* algorithm in three maze environments with different complexity, $25 \times 25$, $35 \times 35$, and $45 \times 45$.
The environments with the promising region prediction are shown in Fig. \ref{fig_test_envs}.
The environments are configured as follows: $256 \times 256$ pixels for environment size, $6$ pixels for RRT* step size.
The involved prediction model is trained on a training dataset with complexity of $31 \times 31$, $33 \times 33$ and $35 \times 35$. 
The maze environments used for the comparison are selected from a test dataset.
During the comparison, L-RRT* ($\alpha=0.5$), L-RRT* ($\alpha=0.8$), and the conventional RRT* are tested 50 times in each environment.
We compare the number of iterations, the number of nodes in the planning process, and the time cost between L-RRT* and conventional RRT*.
The experiment results on finding the optimal path are shown in Fig. \ref{fig_test_result}.
It is noted that we use Python $3.6.10$ to complete the path planning program.

The outcomes illustrate that L-RRT* has a higher sampling efficiency, which results in a faster path planning performance.
L-RRT* achieves much better performance with respect to the three comparison metrics, including the number of iterations, the number of nodes, and the time cost.
Moreover, since the biased-sampler avoids unnecessary search in the dead-end, L-RRT* can perform a more stable path planning.
In general, the experiment results show that our L-RRT* can always outperform the conventional RRT* in terms of the comparison metrics.
The proposed promising region prediction and the biased sampler can dramatically improve the path planning performance. 
It is worth noticing that the prediction model provides insightful heuristic information not only in the maze with trained complexity ($35 \times 35$ in this experiment) but also in the maze with unseen complexity ($25 \times 25$ and $45 \times 45$). 
The experiment results demonstrate the generalization capability of our proposed promising region prediction model. 
This characteristic indicates that the L-RRT* can be easily applied to other different and complex environments and achieve satisfactory performance.
 
\section{Conclusions and Future Work}
\label{sec_conclusion}
In this work, we propose a learning-based path planning algorithm, which directly uses RGB image of the current environment as input to predict efficient heuristic, which guides a RRT* planner to achieve fast path planning. 
The proposed auto-encoder-decoder-like CNN model can generalize well to the unseen environment with unseen complexity. 
A series of simulation experiments have been conducted to show that our proposed method is applicable to different environments and can achieve more efficient sampling and computation than the conventional RRT* algorithm.

For future work, we plan to evaluate the proposed algorithm in real-world applications and further improve its performance.
Another possible avenue is to extend the learning-based path planning method to high-dimensional and complex tasks, where the semantic or natural language information \cite{gopalan2020simultaneously} can be taken into consideration to aid the path planning.
%%%%%%%%%%%%%%%%%%%%%%%%%%%%%%%%%%%%%%%%%%%%%%%%%%%%%%%%%%%%%%%%%%%%%%%%%%%%%%%% 

\addtolength{\textheight}{-7cm}   % This command serves to balance the column lengths

%%%%%%%%%%%%%%%%%%%%%%%%%%%%%%%%%%%%%%%%%%%%%%%%%%%%%%%%%%%%%%%%%%%%%%%%%%%%%%%%

%%%%%%%%%%%%%%%%%%%%%%%%%%%%%%%%%%%%%%%%%%%%%%%%%%%%%%%%%%%%%%%%%%%%%%%%%%%%%%%%
%\section*{Acknowledgement}
%This work is partially supported by Shenzhen Key Laboratory of Robotics Perception and Intelligence, Southern University of Science and Technology, Shenzhen 518055, China, Hong Kong RGC CRF grant C4063-18G, Hong Kong RGC GRF grant \#14200618, Hong Kong ITC ITSP Tier2 grant \#ITS/105/18FP, and Hong Kong ITC MRP grant \#MRP/011/18.
%We also appreciate the great help from the members of Robotics, Perception and Artificial Intelligence Lab in The Chinese University of Hong Kong.

%%%%%%%%%%%%%%%%%%%%%%%%%%%%%%%%%%%%%%%%%%%%%%%%%%%%%%%%%%%%%%%%%%%%%%%%%%%%%%%%
\bibliographystyle{IEEEtran}
\bibliography{reference_JK}

\end{document}